\documentclass[conference]{IEEEtran}
\IEEEoverridecommandlockouts

\usepackage{cite}
\usepackage{amsmath,amssymb,amsfonts}
\usepackage{graphicx}
\usepackage[table]{xcolor}
\usepackage{booktabs}
\usepackage{adjustbox}
\usepackage{array}
\usepackage{comment}
\usepackage{url}

\newcolumntype{L}[1]{>{\raggedright\arraybackslash}m{#1}}
\newcolumntype{C}[1]{>{\centering\arraybackslash}m{#1}}

\newcommand{\method}{\textit{CroCoMAE}}
\newcommand{\variant}{\textit{CroCoMAE$_{NoU}$}}
\newcommand{\decur}{\textit{DeCUR}}
\newcommand{\comm}{\textit{CoMM}}
\newcommand{\croma}{\textit{CROMA}}
\newcommand{\simclr}{\textit{SimCLR}}
\newcommand{\barlow}{\textit{Barlow Twins}}
\newcommand{\mae}{\textit{MAE}}
\newcommand{\simclrC}{\textit{SimCLR-C}}
\newcommand{\barlowC}{\textit{Barlow Twins-C}}

\definecolor{Forest}{HTML}{228B22}
\definecolor{Barren}{HTML}{D2B48C}
\definecolor{Shrubland}{HTML}{90EE90}
\definecolor{Grassland}{HTML}{FFE664}
\definecolor{Cropland}{HTML}{FFA500}
\definecolor{Urban}{HTML}{C83232}
\definecolor{Wetland}{HTML}{40E0D0}
\definecolor{Water}{HTML}{1E90FF}

\definecolor{lightorangefig}{RGB}{255,229,206}
\definecolor{orangefig}{RGB}{255,170,95}
\definecolor{lightbluefig}{RGB}{210,227,240}
\definecolor{lightblue}{RGB}{67,141,192}

\newcommand{\redColor}{red!25}

\newif\ifshowtodos
\showtodostrue
\ifshowtodos
  
  \newcommand{\TODO}[1]{\textbf{\color{red}[TODO: #1]}}
  
\else
  
  \newcommand{\TODO}[1]{}
  
\fi

\definecolor{revcolor}{RGB}{0,140,70}
\newif\ifshowrev
\showrevtrue
\ifshowrev
  
\else
  
\fi

\usepackage[breaklinks,colorlinks,allcolors=blue!60!black]{hyperref}

\begin{document}
\bstctlcite{IEEEexample:BSTcontrol}

\title{\method{}: Cross-modal self-supervised collaborative representation learning for SAR-optical remote sensing data}

% --- Anonymous author block, as required by the IJCNN 2027 double-blind policy.
% \author{\IEEEauthorblockN{Anonymous Authors}}
\author{
\IEEEauthorblockN{Vito Recchia\IEEEauthorrefmark{1}, Giuseppe Guarino\IEEEauthorrefmark{3},
Cássio Fraga Dantas\IEEEauthorrefmark{3}\IEEEauthorrefmark{4},\\
Giuseppina Andresini\IEEEauthorrefmark{1}\IEEEauthorrefmark{2}, and Annalisa Appice\IEEEauthorrefmark{1}\IEEEauthorrefmark{2}, Dino Ienco\IEEEauthorrefmark{3}\IEEEauthorrefmark{4}
}
\IEEEauthorblockA{\IEEEauthorrefmark{1}\textit{Department of Computer Science, University of Bari Aldo Moro}, Bari, Italy}
\IEEEauthorblockA{\IEEEauthorrefmark{2}\textit{Consorzio Interuniversitario Nazionale per l'Informatica (CINI)}, Bari, Italy \\
\IEEEauthorblockA{\IEEEauthorrefmark{3}\textit{INRIA, EVERGREEN, Univ. of Montpellier, Montpellier, France}}
\IEEEauthorblockA{\IEEEauthorrefmark{3}\textit{INRAE, UMR TETIS, Univ. of Montpellier, Montpellier, France}}
\{vito.recchia, giuseppina.andresini, annalisa.appice\}@uniba.it, \\ giuseppe.guarino@inria., {dino.ienco,cassio.fraga-dantas}@inrae.fr}
}
% \author{\IEEEauthorblockN{First Author}
% \IEEEauthorblockA{\textit{Institution} \\
% City, Country \\
% email@address}
% \and
% \IEEEauthorblockN{Second Author}
% \IEEEauthorblockA{\textit{Institution} \\
% City, Country \\
% email@address}}

\maketitle

\begin{abstract}
Remote sensing increasingly captures the Earth's surface through complementary sensors such as SAR and optical, yet ground-truth annotations remain scarce, limiting supervised approaches. Self-supervised learning (SSL) addresses the label scarcity, and cross-modal SSL further exploits the complementarity between co-located sensors, but existing cross-modal frameworks align modalities mostly through global image-level representations or treat them as largely redundant, leaving their fine-grained complementary structure underexploited. We propose \method{}, a cross-modal SSL framework for SAR-optical remote sensing that unifies contrastive learning and masked image modeling within a ViT-based architecture, operating at two complementary levels. At the image level, an intra-modality contrastive loss on the class token yields globally discriminative, augmentation-invariant representations. At the patch level, masked image modeling drives fine-grained per-modality representations, while a cross-modal contrastive objective aligns a shared subspace across SAR and optical patch tokens, leaving the complementary subspace unconstrained to retain modality-specific information. On two SAR-optical benchmarks covering scene classification and semantic segmentation, \method{} improves over uni-modal SSL baselines and outperforms state-of-the-art cross-modal frameworks  in most configurations, with gains in both single-modality and multi-modal evaluation across varying label regimes and the largest improvements on semantic segmentation. Code is available at this~\href{https://anonymous.4open.science/r/CroCoMAE-F0E3/README.md}{link}.
\end{abstract}


\section{Introduction}
\label{sec:intro}
Remote Sensing (RS) is increasingly characterized by the availability of multi-modal sensor data, with multiple satellites and sensors acquiring complementary information over the same region of interest~\cite{yuan2025survey}; SAR and optical sensors are a prime example, complementary in availability as well as information, as SAR operates through cloud and darkness where optical fails. Yet, while the volume of raw multi-sensor RS data grows exponentially, ground truth annotations remain scarce and costly to obtain~\cite{roscher2024better}, %often requiring fresh and abundant expert labeling for each new region or sensor, 
precluding the adoption of standard supervised approaches for RS downstream tasks. Self-supervised learning (SSL) mitigates this imbalance by exploiting abundant unlabeled RS data~\cite{chen2025self}. Existing SSL methods for a single modality broadly fall into two families with complementary strengths: contrastive/distance-based approaches~\cite{Chen2020,Zbontar2021}, which contrast augmented views to learn globally discriminative representations but capture little fine-grained spatial structure, and reconstruction/generative approaches~\cite{He2022,Bachmann2022}, which recover masked input to learn local structure but yield less globally discriminative features. Both are well studied uni-modally in computer vision~\cite{gui2024survey} and RS~\cite{tao2023self}, yet applied to a single modality they cannot exploit the complementarity between co-located sensors. Cross-modal SSL targets this, but existing approaches~\cite{Dufumier2025,Wang2024,fuller2023croma} align the two modalities only through global image-level representations, even when cross-modal contrast is combined with masked reconstruction~\cite{fuller2023croma}, leaving fine-grained, patch-level cross-modal alignment underexplored (Sec.~\ref{sec:related}). To close this gap, in the context of SAR-optical RS data, we propose \method{} (\textit{Cro}ss-modal self-supervised \textit{Co}llaborative \textit{M}asked \textit{A}uto\textit{E}ncoder), a ViT-based framework that unifies contrastive learning and masked image modeling at two complementary levels. At the image level, a within-modality contrastive loss on the class token enforces globally discriminative representations. At the patch level, masked image modeling drives per-modality fine-grained reconstruction, while a cross-modal contrastive term aligns a shared subspace across SAR and optical patch tokens, leaving the unshared subspace free to capture complementary, possibly modality-specific information. Unlike prior work that aligns only global representations, this patch-level objective lets \method{} exploit fine-grained SAR-optical interactions. We evaluate \method{} on two SAR-optical benchmarks covering scene classification and semantic segmentation. Compared to uni-modal SSL baselines and cross-modal frameworks, across varying label regimes and under both linear probing and full fine-tuning, \method{} consistently provides an effective general-purpose initialization. 

In summary, our contributions are: i) \method{}, a cross-modal SSL framework for SAR-optical RS data unifying contrastive learning and masked image modeling at image and patch levels; ii) A patch-level cross-modal contrastive design that aligns a shared subspace across modalities while leaving the unshared subspace free to capture complementary, possibly modality-specific information, enabling fine-grained SAR-optical collaboration beyond global alignment; iii) An extensive evaluation on scene classification and semantic segmentation benchmarks against uni-modal and cross-modal SSL frameworks retrained under the same controlled protocol, showing gains in most configurations across low-data label regimes, particularly on dense prediction. %The learned representations also remain effective under linear probing and single-modality inference, highlighting their potential for flexible use across different sensor-availability configurations.
%\end{itemize}

\section{Related Works}
\label{sec:related}
\noindent\textbf{Uni-Modal Self-Supervised.} Self-supervised learning (SSL) has emerged as a paradigm for learning representations from unlabeled data, eliminating the need for large amounts of costly annotations. Contrastive/distance-based approaches became dominant, with frameworks such as SimCLR~\cite{Chen2020}, MoCo~\cite{He2020} and BYOL~\cite{Grill2020}, and dense variants such as DenseCL~\cite{wang2021densecl}, maximizing agreement between augmented views of the same image. Redundancy-reduction methods, notably Barlow Twins~\cite{Zbontar2021} and VICReg~\cite{Bardes2022}, instead decorrelate feature dimensions to avoid collapse without explicit negative pairs. More recently, masked image modeling regained prominence through Masked Autoencoders (MAE)~\cite{He2022}, showing that reconstruction alone yields rich, transferable representations. Such techniques have been widely adopted in RS: for optical data, e.g., SatMAE~\cite{Cong2022} and SpectralGPT~\cite{Hong2024}, adapting contrastive or masked modeling to multispectral and multi-temporal imagery; and, only very recently, for SAR data, e.g. SARMAE~\cite{liu2026sarmae}.

\noindent\textbf{Multi-Modal Self-Supervised.} A natural extension of Uni-Modal SSL consists in training a single unified model across modalities or sensors. This has led to large multi-modal foundation models, e.g., ImageBind~\cite{Girdhar2023}, which anchors six modalities to a shared image embedding space, and MultiMAE~\cite{Bachmann2022}, which extends masked autoencoding to multi-modal, multi-task reconstruction. At larger scale, several works introduce RS foundation models pretrained on massive external multi-sensor corpora, including %DOFA~\cite{Xiong2024},
SMARTIES~\cite{sumbul2025smarties}, TERRAMIND~\cite{jakubik2025terramind} and AnySat~\cite{astruc2025anysat}. These works target general-purpose pretraining at a scale not directly comparable to the controlled comparison adopted here. Regardless of scale, these approaches learn a common representation across modalities, conflating shared and modality-specific features. This motivates a complementary line of work, discussed next, that instead uses dedicated per-modality encoders with explicit cross-modal alignment, allowing shared and specific information to be represented separately rather than fused. \\
\noindent\textbf{Cross-Modal Self-Supervised.} Purely contrastive cross-modal approaches implicitly assume multiview redundancy: all task-relevant information is shared across modalities, an assumption often violated in practice. \comm{}~\cite{Dufumier2025} addresses this via a contrastive objective in a fused multimodal space, showing empirically that purely cross-modal methods fail to capture unique and synergistic information. \decur{}~\cite{Wang2024} partitions each per-modality encoder's embedding into a common subspace, aligned via Barlow Twins loss, and a unique subspace, explicitly decorrelated across modalities. \croma{}~\cite{fuller2023croma} combines both paradigms: it performs cross-modal contrastive learning on global, average-pooled unimodal representations of masked inputs, with no explicit decomposition, while a separate cross-modal encoder fuses both modalities to jointly reconstruct their masked patches. \decur{} has been evaluated under in-domain training in its original work, whereas \croma{} was pretrained on a large external multi-sensor corpus; since its objectives do not depend on the pretraining corpus, we retrain both methods in-domain under the same controlled protocol as \method{} (Sec.~\ref{sec:expe}). Our work follows a subspace decomposition principle like \decur{}, but differs from all three methods: rather than operating only on global image-level embeddings, \method{} acts also at the level of spatially localised ViT patch tokens within a masked autoencoding framework, leveraging cross-modal contrastive alignment directly within a shared subspace, while reconstruction remains per-modality and independent. This yields a shared/unshared decomposition expressed directly at the patch level, which distinguishes \method{} from prior cross-modal SSL frameworks. Table~\ref{tab:method_comparison} summarizes the design differences across the discussed approaches.

\begin{table}[!ht]
\centering
\footnotesize
\setlength{\tabcolsep}{4pt}
\caption{Comparison of cross-modal SSL designs. Align. columns indicate whether cross-modal (CM) alignment is performed at the global image or patch level.}
\label{tab:method_comparison}
\begin{adjustbox}{max width=\linewidth}
\begin{tabular}{lccccc}
\toprule
 & Global  & Patch  & Partition & MAE & Token\\
 & CM Align. & CM Align. & & & Level \\
\midrule
\croma{}    & \checkmark & --         & --              & \checkmark & Patch\\
\comm{}     & \checkmark & --         & --              & -- & CLS \\
\decur{}    & \checkmark & --         & Shared/Unique   & -- & CLS\\
\method{}  & --         & \checkmark & Shared/Unshared & \checkmark & CLS + Patch\\
\bottomrule
\end{tabular}
\end{adjustbox}
\end{table}

\section{Method}
\label{sec:method}

\begin{figure*}[!ht]
    \centering
    \includegraphics[width=.55\linewidth]{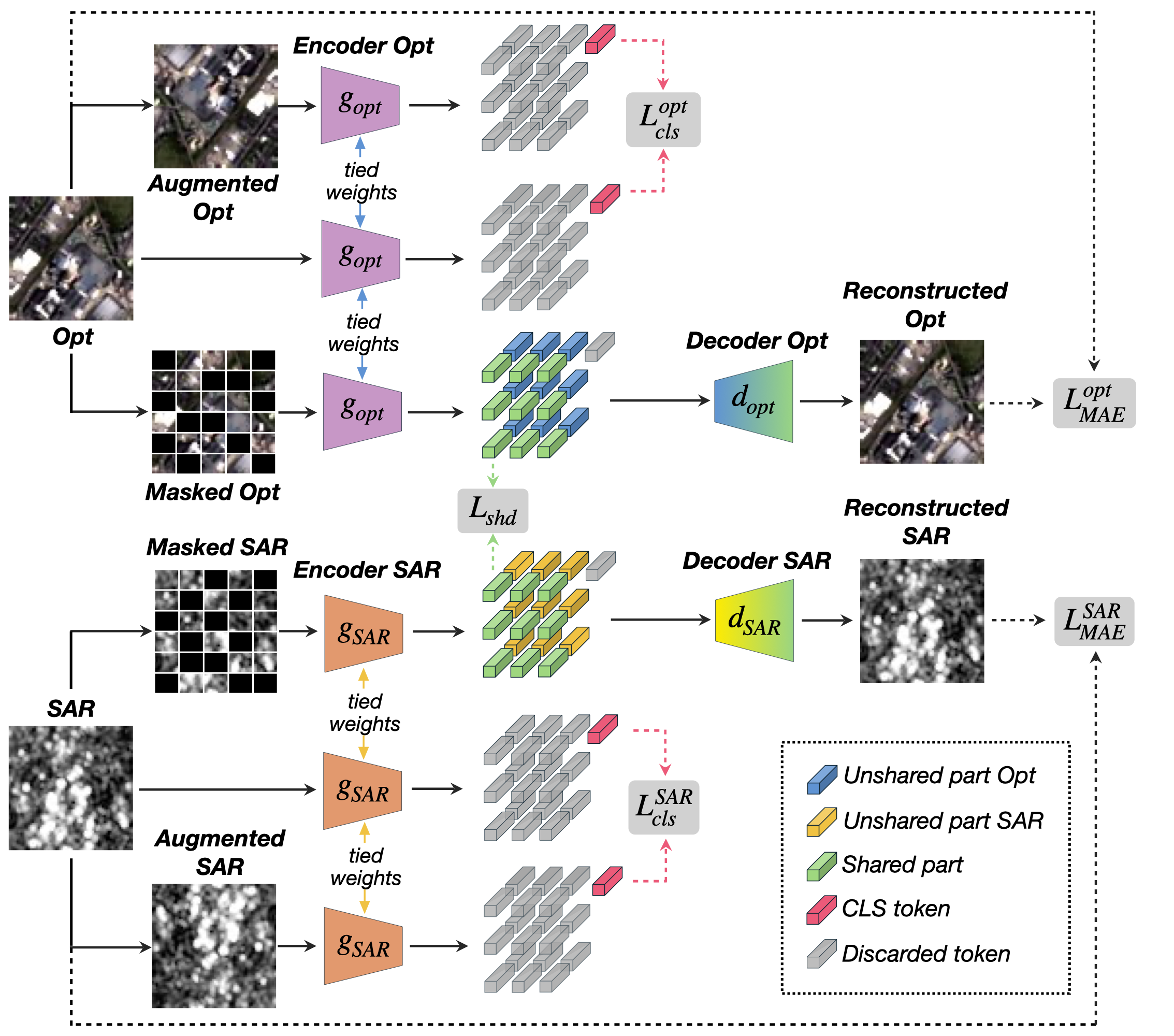}
    \caption{
    Overview of \method{}. For each modality $m$, the encoder $g_m$ processes three inputs: the original image, an augmented view, and a masked version. The resulting tokens are exploited selectively depending on the objective. \textbf{Patch tokens} (green: shared; blue/yellow: unshared) produced from the masked input are used to compute the patch-level losses ($\mathcal{L}_{\mathrm{MAE}}$, $\mathcal{L}_{\mathrm{shd}}$); the CLS token is discarded in this case. Conversely, the \textbf{CLS tokens} (pink) extracted from the original and augmented images are used to compute the image-level contrastive loss $\mathcal{L}_{\mathrm{cls}}$, while the patch tokens are discarded.  Both objectives share the same encoder weights and are optimized jointly.}
    \label{fig:arch_contramae}
\end{figure*}

%\method{} is built on the Vision Transformer (ViT) and combines masked autoencoding with contrastive learning at two levels of representation (Fig.~\ref{fig:arch_contramae}). At the image level, the CLS token is trained with an intra-modal contrastive loss to produce globally discriminative, augmentation-invariant representations. At the patch level, the embedding of each per-modality encoder is split into a shared subspace, aligned across sensors, and an unshared subspace, left free to retain complementary, possibly modality-specific information, while masked reconstruction grounds each encoder in within-modality structure. CLS and patch tokens are used by mutually exclusive objectives, but act on the same encoder weights, updated jointly in a single training stage.

\method{} is built on the Vision Transformer (ViT) architecture and combines masked autoencoding with contrastive learning at two complementary levels of representation. %This design exploits the complementary strengths of the two SSL families discussed in Sec.~\ref{sec:intro}: contrastive learning yields globally discriminative features, while masked image modeling captures fine-grained spatial structure. 
At the image level, the CLS token is trained with an intra-modal contrastive loss to produce globally discriminative, augmentation-invariant representations. At the patch level, the embedding of each per-modality encoder is split into a shared subspace, aligned across sensors, and an unshared subspace, left free to retain complementary, possibly modality-specific information. Unlike prior cross-modal approaches that decompose global image representations, \method{} applies this decomposition to spatially localised patch tokens, enabling fine-grained cross-modal collaboration, while masked reconstruction grounds each encoder in within-modality structure. As illustrated in Fig.~\ref{fig:arch_contramae}, each encoder processes three inputs per image: the original image and an augmented view, whose CLS tokens feed the image-level loss, and a masked version, whose visible patch tokens feed the patch-level losses. The two sets of tokens are thus used by mutually exclusive objectives, which nonetheless act on the same encoder weights, updated jointly in a single training stage.

\subsection{Notation and Definitions}
\label{sec:notation}
Let $m \in \{SAR, Opt\}$ index the two input image modalities. Following standard practice in SAR-optical SSL literature~\cite{fuller2023croma,Wang2024,astruc2025anysat,liu2026sarmae}, we assume that the two image modalities cover exactly the same spatial extent and are co-registered. Given an input image, we denote the input tensors as $\mathbf{X}^{(m)} \in \mathbb{R}^{C_m \times H \times W}$, where $C_m$ is the number of channels specific to modality $m$, and $H \times W$ is the shared spatial resolution. Each modality is processed by a dedicated ViT encoder $f_{\theta_m}$, which partitions the input into $P$ non-overlapping patches, projects each patch into a $K$-dimensional token, and prepends a learnable CLS token. The encoder output is therefore: 
{\small
\[
    \mathbf{Z}^{(m)} = f_{\theta_m}(\mathbf{X}^{(m)}) 
    \;\in\; \mathbb{R}^{(P+1) \times K}
\]}
where $\mathbf{Z}^{(m)}_{\,0,\,:}$ is the CLS token and $\mathbf{Z}^{(m)}_{\,1:,\,:}$ are the patch tokens.
We denote these two components as:
{\small
\[
    \mathbf{h}^{(m)} = \mathbf{z}^{(m)}_0
    \;\in\; \mathbb{R}^{K}
    , 
    \qquad
    \mathbf{P}^{(m)} = \mathbf{z}^{(m)}_{1:} 
    \;\in\; \mathbb{R}^{P \times K}
\]}

The embedding dimension $K$ is split into two equal halves $K_{\mathrm{shd}} = K_{\mathrm{u}} = K/2$, partitioning each row of $\mathbf{P}^{(m)}$ into a shared component $\mathbf{P}^{(m)}_{\mathrm{shd}} \in \mathbb{R}^{P \times K_{\mathrm{shd}}}$ and an unshared component $\mathbf{P}^{(m)}_{\mathrm{u}} \in \mathbb{R}^{P \times K_{\mathrm{u}}}$. The shared subspace is trained to encode information shared across modalities, while the unshared one is unconstrained, allowing complementary, possibly specific information to reside therein. This partition introduces no additional parameters: it is enforced implicitly through the training objectives described later, with no learned gating/routing mechanism. For the patch-level contrastive objective, the patch tokens of all $B$ images in the mini-batch are stacked into a single set $\mathbf{E}^{(m)} \in \mathbb{R}^{N \times K}$ ($N = B \cdot P$), whose rows inherit the same shared/unshared split as $\mathbf{P}^{(m)}$, giving shared and unshared components $\mathbf{E}^{(m)}_{\mathrm{shd}}$ and $\mathbf{E}^{(m)}_{\mathrm{u}}$. 
Finally, since all contrastive objectives share the same form, we define a generic Information Noise-Contrastive Estimation loss $\mathrm{InfoNCE}(\mathbf{U}, \boldsymbol{\ell})$~\cite{Oord2018} on a matrix of $\ell_2$-normalized embeddings $\mathbf{U}$, with positive pairs given by the label vector $\boldsymbol{\ell}$, instantiated per objective below:
{\small
\[
    \mathrm{InfoNCE}(\mathbf{U}, \boldsymbol{\ell}) =
    -\frac{1}{|\mathbf{U}|} \sum_{i} 
    \frac{1}{|\mathcal{P}_i|}
    \sum_{j \in \mathcal{P}_i} 
    \log \frac{
        \exp(\mathbf{U}_i \cdot \mathbf{U}_j )
    }{
        \sum_{k \neq i} \exp(\mathbf{U}_i \cdot \mathbf{U}_k )
    }
\]}
where $\mathcal{P}_i = \{j \neq i : \ell_j = \ell_i\}$ is the set of positives for sample $i$.

\subsection{Image-Level Contrastive Learning via CLS Token}
The CLS token aggregates a global summary of the input image within the ViT encoder. To learn image-level representations that are invariant to radiometric and geometric augmentations, we use a within-modality contrastive objective following SimCLR~\cite{Chen2020}.
For each modality $m$, two views are generated from the input image $\mathbf{X}^{(m)}$: the original image and a randomly augmented counterpart $\tilde{\mathbf{X}}^{(m)} = \mathcal{T}^{(m)}(\mathbf{X}^{(m)})$, where $\mathcal{T}^{(m)}$ is a stochastic augmentation pipeline detailed in Section~\ref{sec:expe}.
The CLS tokens $\mathbf{h}^{(m)}$ and $\tilde{\mathbf{h}}^{(m)}$ are extracted from the original and augmented views, respectively:
{\small
\[
    \mathbf{h}^{(m)} = f_{\theta_m}(\mathbf{X}^{(m)})_0, 
    \qquad
    \tilde{\mathbf{h}}^{(m)} = f_{\theta_m}(\tilde{\mathbf{X}}^{(m)})_0
\]
}
For the contrastive loss, the CLS tokens of both views are concatenated across the mini-batch and $\ell_2$-normalized. Pairs of views from the same image are treated as positives, and all other images in the batch as negatives. Since the loss is computed across all $B$ images in the mini-batch, the label vector $\boldsymbol{\ell}_{\mathrm{img}}$ assigns the same index to both views of the same image. The image-level contrastive loss is:
{\small
\[
    \mathcal{L}_{\mathrm{cls}} = \frac{1}{2} \sum_{m \in \{SAR, Opt\}}
    \mathrm{InfoNCE}\!\left(
        \ell_2\!\left(
            \begin{bmatrix}\mathbf{h}^{(m)} \\ 
            \tilde{\mathbf{h}}^{(m)}\end{bmatrix}
        \right),\; 
        \boldsymbol{\ell}_{\mathrm{img}}
    \right)
\]
}
where $\ell_2(\cdot)$ denotes row-wise $\ell_2$ normalisation, and 
$\mathbf{h}^{(m)}, \tilde{\mathbf{h}}^{(m)} \in \mathbb{R}^{B \times K}$ collect the CLS tokens of the original and augmented views across the mini-batch, respectively.

\subsection{Patch-Level Representation Learning}
Patch tokens encode spatially localised features and constitute the primary substrate for the shared/unshared decomposition. Two complementary objectives act on these tokens.

\noindent \textbf{Reconstruction.} Following MAE~\cite{He2022}, a randomly generated mask at ratio $r$ is applied, leaving $P'=\lfloor (1-r) \cdot P \rfloor$ visible patch tokens per image. For a given multi-modal sample, the same mask is applied to both the SAR and optical images, so that $\mathcal{M}^{(SAR)} = \mathcal{M}^{(Opt)} = \mathcal{M}^{(m)}$ denotes the (shared) set of masked patch positions. Sharing the mask is required by the cross-modal objective introduced below, whose positive pairs are co-registered visible patches. Since patches at the same position in other images of the batch act as negatives, positional information alone cannot solve this objective, which instead requires the shared subspace to encode patch content. Only visible patch tokens are encoded, and a lightweight per-modality decoder reconstructs the pixel values of the masked patches. With per-patch normalisation, the reconstruction loss is:
{\small
\[
    \mathcal{L}_{\mathrm{MAE}} = \frac{1}{2}\sum_{m \in \{SAR, Opt\}}
    \frac{1}{|\mathcal{M}^{(m)}|} \sum_{p \in \mathcal{M}^{(m)}} 
    \left\|\mathbf{X}^{(m)}_p - \mathbf{\hat{X}}^{(m)}_p\right\|_2^2
\]
}
Where $\mathbf{X}^{(m)}_p$ (resp. $\hat{\mathbf{X}}^{(m)}_p$) is the original (resp. reconstructed) masked patch token at position $p \in \mathcal{M}^{(m)}$. 
This loss encourages each encoder to develop rich intra-modal spatial representations independently of the cross-modal objectives, discouraging degenerate solutions that neglect fine-grained spatial structure in favour of cross-modal alignment shortcuts.

\noindent \textbf{Cross-modal shared subspace alignment.}
To align the shared subspace across modalities, we use the InfoNCE loss on the $\ell_2$-normalized shared patch embeddings of both modalities, concatenated into a pool of $2BP'$ tokens in a given batch. Co-registered patch pairs (i.e., patches at the same spatial position from the SAR and optical modalities) act as positives; all other patches serve as negatives. The label vector $\boldsymbol{\ell}_{\mathrm{patch}}$ assigns the same index to the $k$-th patch from both modalities, defining co-registered patch pairs as positives. The cross-modal loss is defined as follows:
{\small
\[
    \mathcal{L}_{\mathrm{shd}} = 
    \mathrm{InfoNCE}\!\left(
        \ell_2\!\left(
            \begin{bmatrix}
                \mathbf{E}^{(SAR)}_{\mathrm{shd}} \\[2pt]
                \mathbf{E}^{(Opt)}_{\mathrm{shd}}
            \end{bmatrix}
        \right),\;
        \boldsymbol{\ell}_{\mathrm{patch}}\
    \right)
\]
}

\subsection{Overall Objective Function}

The final training loss is the unweighted sum of all three terms previously introduced:
{\small
\[
    \mathcal{L}_{\mathrm{total}} = 
    \mathcal{L}_{\mathrm{MAE}} + 
    \mathcal{L}_{\mathrm{shd}} + 
    \mathcal{L}_{\mathrm{cls}}
\]
}
The interplay among the losses is deliberate: $\mathcal{L}_{\mathrm{MAE}}$ grounds each encoder in within-modality spatial structure%, discouraging cross-modal alignment shortcuts
; $\mathcal{L}_{\mathrm{shd}}$ drives patch-level cross-modal alignment in the shared subspace%, allowing modality collaboration
; and $\mathcal{L}_{\mathrm{cls}}$ ensures that the global per-modality image-level representations remain invariant under augmentations. The unweighted formulation avoids loss-balancing hyperparameters, removing the need for per-task tuning.

\section{Experimental Evaluation}
\label{sec:expe}
To evaluate \method{}, we design an experimental study covering two SAR-optical RS benchmarks, for scene classification and semantic segmentation. We compare against both standard uni-modal SSL methods and recent cross-modal approaches. We conduct an ablation study where the contribution of each component of \method{} is assessed, explore the interplay between CLS and patch tokens, and analyse the behaviour of \method{} varying the size of the shared subspace. Finally, we provide a visual inspection of the segmentation maps produced by \method{} and representative competing approaches and a discussion on computational costs.

\noindent\textbf{Datasets.} We evaluate \method{} on two downstream tasks. In both cases, Sentinel-1 SAR and Sentinel-2 multi-spectral optical (Opt) images are used, with SAR images providing two polarization channels and multi-spectral optical images providing 13 spectral bands. \textit{EuroSAT-SAR~\cite{Wang2024a}} (Multi-Class Classification): A SAR extension of the original EuroSAT dataset \cite{Helber2019}, constructed by matching each Sentinel-2 patch with a co-registered Sentinel-1 patch based on geospatial coordinates. The resulting dataset comprises 27,000 SAR-Opt image pairs spanning 10 land-cover classes, with 2,700 samples per class. Each image is 64 $\times$ 64 pixels. \textit{DFC2020~\cite{Robinson2021}} (Semantic Segmentation): Derived from SEN12MS~\cite{Schmitt2019} and released as part of the 2020 IEEE GRSS Data Fusion Contest for weakly supervised semantic segmentation, this dataset provides SAR-Opt image pairs with annotations at two resolutions. We consider the publicly available training and validation sets. From the training set we derive 97,824 SAR-Opt image pairs by tiling the original $256 \times 256$ patches into non-overlapping $64 \times 64$ sub-patches; since the corresponding annotations are at coarse resolution (500m $\times$ 500m), they are discarded. The validation set comprises 15,776 image pairs with high-resolution annotations %(10m $\times$ 10m)
spanning 9 land-cover classes.

\noindent\textbf{Competing Approaches.} We compare \method{} against several competitors spanning both uni-modal and cross-modal SSL strategies. On the uni-modal side, we consider three well-established approaches: \simclr~\cite{Chen2020}, \barlow{}~\cite{Zbontar2021}, and Masked AutoEncoder (\mae)~\cite{He2022}. On the cross-modal side, we consider two state-of-the-art approaches designed for and evaluated on SAR-optical RS data: Decoupling Common and Unique Representations (\decur{})~\cite{Wang2024} and Contrastive Radar-Optical Masked Autoencoders (\croma{})~\cite{fuller2023croma}, all retrained in-domain under the same controlled protocol, following the original available implementations and recommendations, adapted to the shared ViT-Small backbone.
We further include \simclrC{} and \barlowC~\cite{Wang2024}, which extend their respective uni-modal counterparts to the cross-modal setting by treating the corresponding image from the other modality (SAR or Opt) as an augmentation of the input. 
We additionally include \variant{} as an internal variant of our framework, in which the $\mathcal{L}_{\mathrm{shd}}$ loss is applied directly to the full patch embedding, without partitioning it into shared and unshared subspaces. For all SSL methods we use a ViT-Small backbone~\cite{touvron2021training}. We additionally report two supervised baselines using ViT-Small and ResNet-50. For the multi-class classification task (\textit{EuroSAT-SAR}), all backbones are combined with a single-layer classifier. For semantic segmentation (\textit{DFC2020}), all ViT-based backbones are paired with a lightweight PSPNet-style pyramid pooling head~\cite{zhao2017pyramid} applied to the final-layer patch tokens, while ResNet-50 is used within a UNet~\cite{Ronneberger_2015_MICCAI} architecture baseline.

\noindent\textbf{Experimental Settings.} All SSL approaches share the same data, backbone and fine-tuning setup, and are pre-trained with AdamW~\cite{Loshchilov2019} (weight decay 0.05) for 260,000 steps (batch size 512), with a maximum learning rate of $1.5\times10^{-4}$, a linear warm-up of 20 epochs, followed by cosine annealing with warm restarts.
We evaluate single-modal (SAR, Opt) and multi-modal (MM) settings, the latter obtained by concatenating per-modality embeddings before the classification head. Fine-tuning uses two labeled regimes (1\% / 10\% of data, with the remaining 99\% / 90\% held out for testing) under two strategies: Linear Probing (LP, frozen encoder) and Full Fine-Tuning (Full-FT, all weights updated), for 300 epochs with batch size 64. For \textit{DFC2020}, LP denotes a frozen encoder with a trainable segmentation head. 

We follow standard SSL augmentations~\cite{Grill2020} adapted to multi-channel inputs (resized crop with scale 0.2--1, horizontal/vertical flip, Gaussian blur, brightness/contrast jitter, and a grayscale conversion replacing all channels with their mean, applicable to any number of bands), and a masking ratio $r=75\%$ for all MAE-based methods~\cite{He2022}. Encoders use $8\times8$-pixel patches; the per-modality decoders of \method{} have 2 layers (128-d, 8 heads), 
and the CLS projection head, discarded after pre-training, is a two-layer MLP (512--128). For contrastive objectives, we use a single temperature $\tau=1$ for all methods, to avoid per-method tuning, except for \croma{}, whose temperature is learnable (initialised at 0.07) as in its original design. Following~\cite{Wang2024}, we use the ViT class token for classification heads and patch tokens for segmentation; \croma{}, which has no class token in its original design, is evaluated with average-pooled patch tokens for classification. Performance is measured via weighted F1 on \textit{EuroSAT-SAR} and mIoU on \textit{DFC2020}, averaged over five runs. Using a protocol identical across methods, for \textit{EuroSAT-SAR} we adopt a transductive setup where pre-training and fine-tuning share the same images (labels used only during fine-tuning); for \textit{DFC2020}, we pre-train on the training split and fine-tune/evaluate on the disjoint validation split. This inductive protocol, in which evaluation images are never seen during pre-training, complements the transductive \textit{EuroSAT-SAR} setting.
Experiments run on an HPC facility with NVIDIA H100 SXM5 GPUs (80\,GB HBM3).

\subsection{Results}
Table~\ref{tab:eurosat} and Table~\ref{tab:dfc2020} report the results on the \textit{EuroSAT-SAR} and \textit{DFC2020} benchmarks, respectively. Standard deviations over the 5 runs are generally below 1.5 points at 1\% labels and below 0.5 at 10\%; the gains of \method{} over the strongest competitors exceed one standard deviation in most configurations. Overall, \method{} achieves the best performance in the majority of the configurations, demonstrating its effectiveness and generality. Across benchmarks and label regimes, \method{}'s Linear Probing performance closely matches Full Fine-Tuning, showing that the pretrained representations are already effective without costly task-specific adaptation. In contrast, several competing methods, most notably \barlow{} on \textit{EuroSAT-SAR} and \decur{} on \textit{DFC2020}, exhibit substantially larger LP/Full-FT discrepancies, indicating that their representations require more extensive fine-tuning to become useful for downstream tasks. This property is particularly valuable in low-data regimes, where the cost of fine-tuning itself may be a practical constraint.

On \textit{EuroSAT-SAR}, \method{} consistently surpasses all supervised baselines, with gains ranging from about 4 points (Opt, 10\% labels) to about 20 points (MM, 1\% labels). Among competing SSL approaches, \simclrC{} and \barlowC{} achieve the highest SAR LP scores at 1\% labels (82.4\% and 82.5\%, respectively). However, this performance on the SAR modality comes at the cost of a marked degradation on the optical modality (84.6\% and 83.2\% vs.\ 96.5\% for \method{}), a trend that is even more pronounced for \croma{} (73.1\% Opt LP, the lowest among cross-modal methods). This suggests that cross-modal strategies aligning the entire global representation tend to favour the lower-performing modality (SAR) at the expense of the higher-performing one (optical), on this benchmark. \method{} avoids this imbalance through its within-modality objectives and consistently achieves the best results in the multi-modal evaluation setting.

\begin{table*}[!t]
\caption{Results on EuroSAT-SAR. We report LP / Full-FT Weighted F1-Score (\%) averaged over 5 runs, for single modal (SAR, Opt) and multi-modal (MM) settings, with 1\% and 10\% labeled data. \colorbox{\redColor}{Red} highlights results below the best supervised baseline. Bold indicates the best result per column. \label{tab:eurosat}}
\centering
\begin{adjustbox}{max width=\linewidth}
\begin{tabular}{l|l|ccc|ccc}
\hline
\multicolumn{8}{c}{EuroSAT-SAR} \\ \hline
& & \multicolumn{3}{c|}{1\%} & \multicolumn{3}{c}{10\%} \\ \hline
& & SAR      & Opt      & MM  & SAR     & Opt & MM \\ \hline
Sup. & ViT &  57.8  & 75.0    &  76.5   & 69.7 & 91.8 & 87.7 \\
& ResNet-50 &  58.2  &  78.7  &  60.1 & 73.1 & 93.7 & 85.4 \\ \hline
Uni- & MAE &  67.3 / 67.2  &  89.9 / 85.3  &  91.0 / 89.5 & 78.0 / 74.0 & 95.4 / 94.8 & 95.9 / 95.1    \\
Modal & SimCLR &\colorbox{\redColor}{48.2} / \colorbox{\redColor}{48.6}  & \colorbox{\redColor}{65.0} / \colorbox{\redColor}{73.1} &  \colorbox{\redColor}{67.7} / \colorbox{\redColor}{71.8}  & \colorbox{\redColor}{55.5} / \colorbox{\redColor}{61.6} & \colorbox{\redColor}{80.2} / \colorbox{\redColor}{88.3} &  \colorbox{\redColor}{82.2} / \colorbox{\redColor}{85.9}   \\
& BarlowT & \colorbox{\redColor}{24.4} / 62.5 & 93.6 / 90.8 &  94.1 / 91.6 & \colorbox{\redColor}{29.4} / \colorbox{\redColor}{70.6} & 96.0 / 95.8 &  96.2 / 96.1    \\
\hline
Cross- & SimCLR-C  &  82.4 / 71.4 &  84.6 / 88.4 &  84.0 / 87.0 & 82.8 / 74.6 & 87.0 / 93.9 & 86.9 / 91.3   \\
Modal& BarlowT-C &  \textbf{82.5} / 69.5  &  83.2 / 86.2 &   79.5 / 87.6 & \textbf{88.6} / \colorbox{\redColor}{72.1} & \colorbox{\redColor}{89.8} / \colorbox{\redColor}{93.1} &  89.9 / 92.8   \\
& DeCUR & 75.9 / 71.6  & 94.1 / 92.7 &   92.7 / 94.2  & 79.2 / 75.8 & 96.3 / 96.6 &   96.3 / 97.0  \\
& CROMA & 72.3 / 68.6 & \colorbox{\redColor}{73.1} / 84.6 & 83.3 / 88.7 & 81.3 / 76.2 & \colorbox{\redColor}{87.8} / \colorbox{\redColor}{92.6} & 92.2 / 94.8 \\ \hline
& \variant{} & 72.2 / \textbf{79.7} & \textbf{96.5} / 91.8 & 95.0 / 90.9 & 79.4 / \textbf{81.1} & \textbf{97.8} / 97.0 & \textbf{97.8} / 97.0  \\
& \method{} & 77.1 / 78.4 & \textbf{96.5} / \textbf{96.0} & \textbf{96.4} / \textbf{95.3} & 82.7 / 80.9 & \textbf{97.8} / \textbf{97.6} & \textbf{97.8} / \textbf{97.7} \\
\hline
\end{tabular}
\end{adjustbox}
\end{table*}

\textit{DFC2020} provides an informative assessment of the quality of the fine-grained spatial representations produced by the proposed pre-training strategy. \method{} is the only strategy that consistently outperforms the supervised baselines across all considered configurations. Notably, this holds even under LP, where encoder weights are frozen: at 10\% labels, \method{} reaches 58.1\% mIoU on the SAR modality and 68.4\% mIoU in the multi-modal setting, exceeding the supervised ViT baseline (32.9\% and 68.2\%, respectively). The obtained results highlight that \method{}'s patch-level representation learning design, which couples intra-modal masked image modeling and patch-level cross-modal alignment, appears well adapted for tasks such as semantic segmentation.

\begin{table*}[!t]
\caption{Results on DFC2020. We report LP / Full-FT mIoU (\%) averaged over 5 runs, for single modal (SAR, Opt) and multi-modal (MM) settings, with 1\% and 10\% labeled data. \colorbox{\redColor}{Red} highlights results below the best supervised baseline. Bold indicates the best result per column.\label{tab:dfc2020}}
\centering
\begin{adjustbox}{max width=\linewidth}
\begin{tabular}{l|l|ccc|ccc}
\hline
\multicolumn{8}{c}{DFC2020} \\ \hline
& & \multicolumn{3}{c|}{1\%} & \multicolumn{3}{c}{10\%} \\ \hline
& & SAR      & Opt      & MM  & SAR     & Opt & MM \\ \hline
Sup. & ViT-PPM & 25.2 &  49.3   & 59.2  & 32.9 & 56.1 & 68.2 \\
& UNet & 29.2 & 48.3 & 54.1  & 33.4 & 54.7 & 67.1 \\ \hline
Uni- & MAE &  39.9 / 39.3  &  50.7 / \colorbox{\redColor}{48.8}   &  \colorbox{\redColor}{51.7} / \colorbox{\redColor}{51.1} & 47.6 / 49.4  &  59.2 / 60.9   &  \colorbox{\redColor}{60.9} / \colorbox{\redColor}{65.7}   \\
Modal & SimCLR &  46.6 / 47.1 &   56.1 / 55.3    &  \textbf{60.4} / 60.8  &  55.4 / 56.4  & 65.8 / 66.4 &  \colorbox{\redColor}{67.7} / 69.0    \\
& BarlowT & 40.5 / 39.2   &  49.9 / 51.4    &  \colorbox{\redColor}{52.0} / \colorbox{\redColor}{53.3}   &  48.8 / 50.3 &  60.1 / 66.4   &  \colorbox{\redColor}{61.9} / \colorbox{\redColor}{66.5}   \\
\hline
Cross- & SimCLR-C  &
45.4  / 45.7    &  55.4 / 56.2   &  \colorbox{\redColor}{57.6} / 59.3   &  52.9 / 53.5   &  65.2 / 66.0   &  \colorbox{\redColor}{66.2} / 68.7 \\
Modal& BarlowT-C & 40.9 / 40.7    &  54.9 / 55.8   &  \colorbox{\redColor}{54.0} / \colorbox{\redColor}{57.8}  &  45.5 / 47.6   &  63.9 / 65.8   &  \colorbox{\redColor}{63.8} / 68.2    \\
& DeCUR &  38.6 / 45.9   &  \colorbox{\redColor}{48.4} / 55.0   &   \colorbox{\redColor}{47.2} / \colorbox{\redColor}{59.1}    & 46.4 / 56.7  &  57.0 / 67.5  &  \colorbox{\redColor}{57.6} / 69.2  \\
& CROMA & 45.0 / 44.3 & 51.1 / 51.2 & \colorbox{\redColor}{59.1} / \colorbox{\redColor}{59.0} & 55.1 / 55.1 & 63.5 / 63.5 & \colorbox{\redColor}{68.0} / 69.4 \\ \hline
& \variant{} & 48.0 / 48.7 & 55.8 / 56.5 & \colorbox{\redColor}{58.4} / 60.5 & 56.0 / 58.2 & 64.9 / 66.8  & \colorbox{\redColor}{66.6} / 69.7 \\
& \method{} & \textbf{51.4} / \textbf{51.4} &  \textbf{58.2} / \textbf{58.1} &  60.1 / \textbf{60.9} & \textbf{58.1} / \textbf{58.8} & \textbf{67.5} / \textbf{67.9} &  \textbf{68.4} / \textbf{69.8} \\
\hline
\end{tabular}
\end{adjustbox}
\end{table*}

To investigate the contribution of the explicit subspace partitioning, we compare \method{} against its ablated variant \variant{}, which applies the cross-modal contrastive loss over the full patch embedding. On \textit{EuroSAT-SAR}, \method{} matches or outperforms \variant{} in most configurations, with the largest gains at 1\% labels (e.g.\ +4.9 for SAR under LP and +4.4 for MM under Full-FT), while at 10\% labels the two variants mostly fall within a narrow margin. On \textit{DFC2020}, where patch tokens are directly exploited for dense prediction, \method{} outperforms \variant{} across both label regimes and all modalities, with the clearest margins under linear probing (+1.7 to +3.4 mIoU). This provides evidence that explicitly structuring the ViT patch-level embedding into shared/unshared subspaces benefits fine-grained spatial discrimination.

\noindent\textbf{Ablation Analysis.} Combining reconstruction and contrastive objectives generally outperforms either alone (Tables~\ref{tab:eurosat} and~\ref{tab:dfc2020}), motivating a closer look at their interplay. Table~\ref{tab:ablation_losses} examines the impact of each loss term on \textit{EuroSAT-SAR}. The last row corresponds to \method{} which achieves the highest average performance on Opt and MM (e.g., 96.5/96.0 and 96.4/95.3 at 1\%), showing that all terms provide complementary empirical benefits. Ablating $\mathcal{L}_{\mathrm{cls}}$ yields the strongest SAR performance (82.7/80.8 at 1\%, 84.7/82.0 at 10\%) but degrades Opt and MM, indicating $\mathcal{L}_{\mathrm{cls}}$ is key to discriminative Opt representations, at some cost to SAR-specific discriminability. $\mathcal{L}_{\mathrm{shd}}$'s contribution is mainly visible under LP: the full model improves consistently over the configuration without $\mathcal{L}_{\mathrm{shd}}$ (69.5$\to$77.1 for SAR, 96.3$\to$96.5 for Opt at 1\%), though this gap narrows under Full-FT, suggesting that, on this benchmark, $\mathcal{L}_{\mathrm{shd}}$ improves the quality of pretrained representations, while Full-FT is able to partially compensate for its absence. Overall, the two contrastive terms play complementary roles: $\mathcal{L}_{\mathrm{cls}}$ for global discriminability, $\mathcal{L}_{\mathrm{shd}}$ for frozen-representation quality, alongside the spatial grounding from $\mathcal{L}_{\mathrm{MAE}}$.

\begin{table}[!t]
\caption{\method{}'s loss terms ablation on EuroSAT-SAR. We report LP / Full-FT Weighted F1-Score (\%) averaged over 5 runs, for SAR, Opt and multi-modal (MM) settings, with 1\% and 10\% labeled data. Bold indicates the best result per column. \label{tab:ablation_losses}}
\centering
\begin{adjustbox}{max width=\linewidth}
\begin{tabular}{ccc|ccc|ccc}
\hline
\multicolumn{9}{c}{EuroSAT-SAR} \\ \hline
\multicolumn{3}{c|}{} & \multicolumn{3}{c|}{1\%} & \multicolumn{3}{c}{10\%} \\ \hline
$\mathcal{L}_{\mathrm{MAE}}$ & $\mathcal{L}_{\mathrm{shd}}$ & $\mathcal{L}_{\mathrm{cls}}$ & SAR      & Opt      & MM  & SAR     & Opt & MM \\ \hline
&  & \checkmark & 48.2 / 48.6  & 65.0 / 73.1 &  67.7 / 71.8  & 55.5 / 61.6 & 80.2 / 88.3 &  82.2 / 85.9   \\
\checkmark & & &  67.3 / 67.2  &  89.9 / 85.3  &  91.0 / 89.5 & 78.0 / 74.0 & 95.4 / 94.8 & 95.9 / 95.1 \\
& \checkmark & \checkmark & 73.7 / 79.6 & 95.7 / 94.8 & 95.6 / 94.6 & 80.4 / 81.2 & 97.0 / 96.8 & 97.1 / 97.0 \\
\checkmark & \checkmark & & \textbf{82.7} / \textbf{80.8} & 90.0 / 90.4 & 88.9 / 90.4 & \textbf{84.7} / \textbf{82.0} & 92.8 / 95.0 & 92.9 / 95.0  \\
\checkmark & & \checkmark & 69.5 / 70.5 &  96.3 / \textbf{96.3} & 96.1 / \textbf{96.3} & 76.4 / 78.3 & 97.6 / \textbf{97.7} & 97.7 / \textbf{98.0} \\
\checkmark & \checkmark & \checkmark & 77.1 / 78.4 & \textbf{96.5} / 96.0 & \textbf{96.4} / 95.3 & 82.7 / 80.9 & \textbf{97.8} / 97.6 & \textbf{97.8} / 97.7 \\ \hline
\end{tabular}
\end{adjustbox}
\end{table}

\noindent\textbf{Analysis of CLS/Patch tokens interplay.} Table~\ref{tab:ablation_token} reports the LP performance on \textit{EuroSAT-SAR} of different token representations extracted from the frozen \method{} encoders; we restrict this analysis to LP, as it directly probes the information encoded in the pretrained representations, which Full-FT would partly override. First, the pooled patch tokens are globally discriminative: they outperform the class token on SAR (81.8 vs.\ 77.1 at 1\%, 84.6 vs.\ 82.7 at 10\%), while the class token remains slightly better in the optical and multi-modal settings (e.g., 96.5 vs.\ 95.3 for Opt at 1\%), indicating that the two levels emphasise different information. Their concatenation closely tracks the better of the two in every setting, without requiring a choice between them. Second, the two subspaces play distinct roles. The shared subspace yields almost identical scores for SAR and optical (80.6 vs.\ 80.3 at 1\%, 81.9 for both at 10\%), and combining the two modalities brings little gain (81.5 and 83.0 for MM), indicating that it encodes the same, cross-modally aligned information in both sensors. Its accuracy of over 80\% further confirms that it captures scene content rather than positional cues. In contrast, the unshared subspace retains most of the optical discriminative power (92.1 vs.\ 80.3 for the shared subspace at 1\%), and combining the unshared parts of both modalities clearly improves over each of them (95.1 for MM vs.\ 92.1 for Opt and 78.8 for SAR at 1\%), showing that they capture complementary, modality-specific information; for SAR, in contrast, neither subspace dominates (80.6 vs.\ 78.8 at 1\%, 81.9 vs.\ 84.3 at 10\% for shared vs.\ unshared), suggesting that its discriminative information is spread across both. Finally, for single-modal inference the full patch embedding outperforms either subspace alone, confirming that both contribute to the learned representation. %A direct measurement on the frozen encoders confirms this decomposition: co-registered SAR and optical patches are nearly identical in the shared subspace (mean cosine similarity of 0.96 and 0.99 on \textit{DFC2020} and \textit{EuroSAT-SAR}, vs.\ 0.00 for random pairs; linear CKA~\cite{kornblith2019similarity} of 0.87 and 0.93), whereas in the unshared subspace they are no more similar than random pairs (cosine similarity of 0.01 and $-$0.01; CKA of 0.30 and 0.18).
A direct measurement on the frozen encoders confirms that this complementary information is modality-specific rather than redundant: co-registered SAR and optical patches are nearly identical in the shared subspace (mean cosine similarity of 0.96 and 0.99 on \textit{DFC2020} and \textit{EuroSAT-SAR}, vs.\ 0.00 for random pairs; linear CKA (Centered Kernel Alignment)~\cite{kornblith2019similarity} of 0.87 and 0.93), whereas in the unshared subspace they are no more similar than random pairs (cos. sim. of 0.01 and $-$0.01; CKA of 0.30 and 0.18).

\begin{table}[!t]
\caption{\method{}'s analysis of the token representation on \textit{EuroSAT-SAR}. We report LP Weighted F1-Score (\%) averaged over 5 runs, for SAR, Opt and MM settings with 1\% and 10\% labeled data. \textit{CLS}: class token; \textit{Patch}: average-pooled patch tokens; \textit{Patch(S)} / \textit{Patch(U)}: shared / unshared subspace of the pooled patch tokens (half the embedding dimension); \textit{CLS+Patch}: concatenation of class and pooled patch tokens. Bold indicates the best result per column.\label{tab:ablation_token}}
\centering
\footnotesize
\setlength{\tabcolsep}{4pt}
\begin{tabular}{l|ccc|ccc}
\hline
& \multicolumn{3}{c|}{1\%} & \multicolumn{3}{c}{10\%} \\ \hline
& SAR & Opt & MM & SAR & Opt & MM \\ \hline
\textit{CLS}       & 77.1 & \textbf{96.5} & \textbf{96.4} & 82.7 & \textbf{97.8} & \textbf{97.8} \\
\textit{Patch}     & \textbf{81.8} & 95.3 & 94.3 & \textbf{84.6} & 97.0 & 97.1 \\
\textit{Patch(S)}  & 80.6 & 80.3 & 81.5 & 81.9 & 81.9 & 83.0 \\
\textit{Patch(U)}  & 78.8 & 92.1 & 95.1 & 84.3 & 96.6 & 97.2 \\
\textit{CLS+Patch} & 80.5 & 96.3 & 96.1 & \textbf{84.6} & 97.6 & 97.7 \\ \hline
\end{tabular}
\end{table}

\begin{figure}[!t]
    \centering
    \includegraphics[width=.86\linewidth]{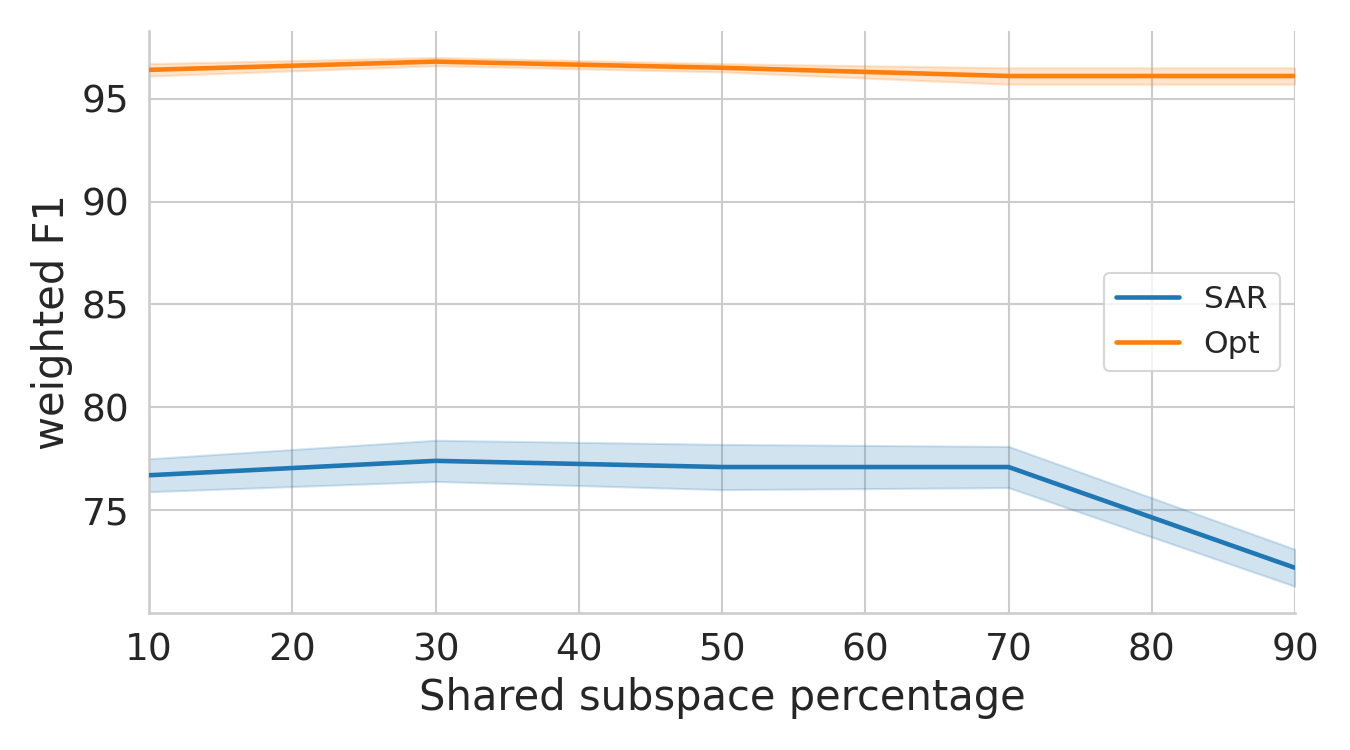}\\[-1pt]
    {\footnotesize (a) \textit{EuroSAT-SAR} (weighted F1)}\\[4pt]
    \includegraphics[width=.86\linewidth]{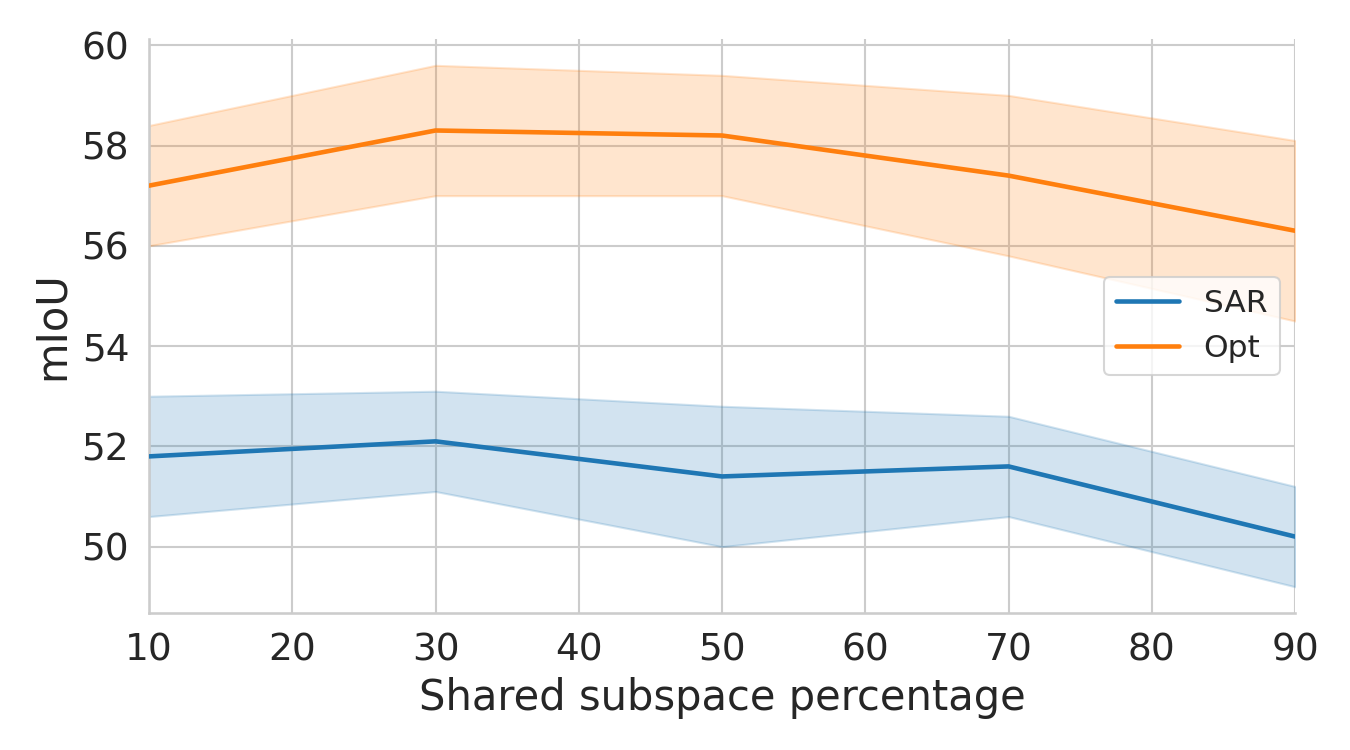}\\[-1pt]
    {\footnotesize (b) \textit{DFC2020} (mIoU)}
    \caption{\method{}'s analysis of the proportion of the patch token embedding allocated to the shared subspace, on (a) \textit{EuroSAT-SAR} and (b) \textit{DFC2020}. We report results averaged over 5 runs, under LP with 1\% labeled data, for \textcolor{lightblue}{SAR} and \textcolor{orangefig}{Opt} modalities. The solid line represents the average and the shaded band the standard deviation.}
    \label{fig:sensitivity_shared}
\end{figure}

\noindent\textbf{Analysis of the Shared/Unshared Subspace Partition.} We investigate the sensitivity of \method{} to the proportion of the patch token embedding allocated to the shared subspace, varying it from 10\% to 90\% in steps of 20\%. Fig.~\ref{fig:sensitivity_shared} reports the weighted F1 score on \textit{EuroSAT-SAR} and the mIoU on \textit{DFC2020} under LP with 1\% labeled data, for the SAR and optical modalities. Two consistent trends emerge on both benchmarks. First, performance is largely insensitive to the partition over the 10--70\% range: variations stay within about one standard deviation (at most 0.7 points on \textit{EuroSAT-SAR} and 1.1 mIoU on \textit{DFC2020}), with a slight optimum around 30\% on both datasets. Second, allocating most of the embedding to the shared subspace consistently degrades performance: from 70\% to 90\% shared, SAR drops by about 5 points on \textit{EuroSAT-SAR} (77.1$\to$72.2) and by about 1.4 mIoU on \textit{DFC2020} (51.6$\to$50.2), and the optical modality loses about 1.1 mIoU on \textit{DFC2020} (57.4$\to$56.3), while the optical F1 on \textit{EuroSAT-SAR}, already close to saturation ($\sim$96\%), is only marginally affected. These results indicate that cross-modal alignment does not require a large shared capacity to be beneficial, whereas leaving too little capacity to the unshared subspace prevents each encoder from retaining complementary, modality-specific information; this is consistent with the advantage of \method{} over \variant{} (Tables~\ref{tab:eurosat} and~\ref{tab:dfc2020}), which corresponds to aligning the full embedding. The default 50/50 split lies within the stable regime on both benchmarks, at most 0.7 points from the best configuration, supporting it as a reasonable, dataset-agnostic choice rather than a per-benchmark optimum.

\noindent\textbf{Qualitative assessment.} Fig.~\ref{fig:qualitative} compares \method{}, \decur{} and the supervised ViT-PPM baseline on \textit{DFC2020} under the multi-modal LP setting with 10\% labeled data. Although all predictions are spatially coarse, as expected with a frozen encoder, \method{} shows sharper class boundaries and better captures intra-scene heterogeneity: it separates compact Barren/Shrubland/Grassland regions (1st row), delineates Urban and Forest classes more accurately (2nd/3rd rows), and is the only method that preserves the continuous geometry of a river (4th row), suggesting that its patch-level objective encourages spatially fine-grained representations.
%Fig.~\ref{fig:qualitative} compares \method{}, \decur{} and the supervised ViT-PPM baseline on \textit{DFC2020} semantic segmentation under the multi-modal LP setting with 10\% labeled data. All methods produce spatially coarse predictions relative to the ground truth, a known limitation when the encoder is frozen. \method{} nonetheless consistently shows sharper class boundaries and better captures intra-scene heterogeneity: it correctly resolves compact Barren/Shrubland/Grassland regions that \decur{} fails to separate (1st row), more accurately delineates Urban and Forest classes where competitors produce fragmented boundaries or overestimate Shrubland due to spectral similarity with vegetation (2nd/3rd rows), and is the only method that preserves the elongated, spatially continuous geometry of a river that all competitors either fragment or miss entirely (4th row). These results suggest that the patch-level training objective of \method{} successfully encourages spatially fine-grained representations.

\newcommand{\swatch}[1]{%
    \fcolorbox{black!40}{#1}{\rule{0pt}{3pt}\hspace{3pt}}%
}
\newcommand{\sizecell}{0.069\textwidth}
\begin{figure}[!t]
    \centering
    \setlength{\tabcolsep}{1.5pt}
    \small
    \begin{tabular}{cccccc}
        \toprule
        SAR & Opt & GT & ViT-PPM & DeCUR & \method{} \\
        \midrule
        \includegraphics[width=\sizecell]{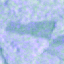} &
        \includegraphics[width=\sizecell]{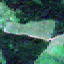} &
        \includegraphics[width=\sizecell]{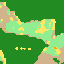} &
        \includegraphics[width=\sizecell]{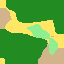} &
        \includegraphics[width=\sizecell]{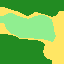} &
        \includegraphics[width=\sizecell]{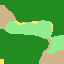} \\[2pt]
        \includegraphics[width=\sizecell]{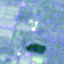} &
        \includegraphics[width=\sizecell]{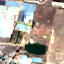} &
        \includegraphics[width=\sizecell]{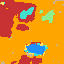} &
        \includegraphics[width=\sizecell]{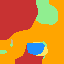} &
        \includegraphics[width=\sizecell]{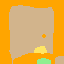} &
        \includegraphics[width=\sizecell]{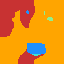} \\[2pt]
        \includegraphics[width=\sizecell]{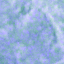} &
        \includegraphics[width=\sizecell]{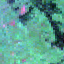} &
        \includegraphics[width=\sizecell]{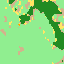} &
        \includegraphics[width=\sizecell]{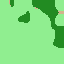} &
        \includegraphics[width=\sizecell]{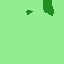} &
        \includegraphics[width=\sizecell]{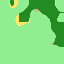} \\[2pt]
        \includegraphics[width=\sizecell]{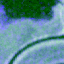} &
        \includegraphics[width=\sizecell]{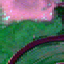} &
        \includegraphics[width=\sizecell]{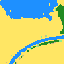} &
        \includegraphics[width=\sizecell]{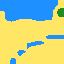} &
        \includegraphics[width=\sizecell]{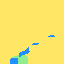} &
        \includegraphics[width=\sizecell]{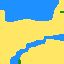} \\
        \bottomrule
    \end{tabular}

    \vspace{3pt}
    {\footnotesize
    \setlength\tabcolsep{.7pt}
    \renewcommand{\arraystretch}{1.2}
    \begin{tabular}{cl @{\hspace{8pt}} cl @{\hspace{8pt}} cl @{\hspace{8pt}} cl }
        \swatch{Forest}    & Forest    &
        \swatch{Barren}    & Barren    &
        \swatch{Shrubland} & Shrubland &
        \swatch{Grassland} & Grassland \\
        \swatch{Cropland}  & Croplands &
        \swatch{Urban}     & Urban     &
        \swatch{Wetland}   & Wetland   &
        \swatch{Water}     & Water     \\
    \end{tabular}
    }
    \caption{Qualitative comparison on \textit{DFC2020} semantic segmentation under the multi-modal LP setting with 10\% labeled data.}
    \label{fig:qualitative}
\end{figure}

\noindent \textbf{Computational Cost.} We restrict this comparison to the cross-modal frameworks \decur{} and \croma{}, excluding uni-modal-extended baselines. Both share \method{}'s two-encoder, ViT-Small-per-modality architecture, totaling approximately 43M parameters per method; the shared/unshared partition in \method{} introduces no additional parameters. The only overhead specific to \method{} comes from its two lightweight per-modality decoders (1.2M parameters, $\sim$3\% of the backbone), whereas \croma{}'s cross-modal fusion encoder and joint decoder add $\sim$18M parameters; in both cases these modules are discarded after pre-training and add no cost at fine-tuning or inference. On \textit{DFC2020}, pre-training takes 34 GPU-hours for \method{} and 32 for \decur{}, a similar trend is observed on \textit{EuroSAT-SAR}. \croma{} is cheaper (10 GPU-hours), since its contrastive and reconstruction objectives share a single forward pass on masked inputs; the additional cost of \method{} comes from the full-image forward passes required by $\mathcal{L}_{\mathrm{cls}}$ and from the dense patch-level contrast of $\mathcal{L}_{\mathrm{shd}}$, the components that drive its gains on dense prediction.

\section{Conclusion}
\label{sec:conclu}
We presented \method{}, a novel cross-modal SSL framework for SAR-optical RS data that unifies  contrastive learning and masked image modeling,  operating at two complementary levels: image and patch. At the patch level, the per-modality embedding is split into a shared subspace, aligned across sensors, and an unshared subspace, left free to retain complementary, possibly modality-specific information. Across scene classification and semantic segmentation benchmarks, \method{} achieves the strongest overall performance across low-data label regimes, training strategies and sensor-availability configurations, with the largest gains on dense prediction tasks. Future work includes extending \method{} to non-RS modality pairs (e.g., RGB-Depth/Thermal) and exploring principled mechanisms for effectively combining image- and patch-level representations.

%\noindent\textbf{Limitations.} Our comparison is restricted to methods trained in-domain under the same controlled protocol; large-scale foundation models pretrained on massive external corpora fall outside this scope and are left for future comparison. \method{} is currently designed for two modalities, and extending the shared/unshared subspace decomposition to a higher number remains challenging. The unshared subspace is also left unconstrained by design; future work could investigate more structured shared/unshared partitions. 

\section*{Acknowledgment}
The authors used Claude (Claude Opus 5.5, Anthropic)~\cite{anthropic2026claude} to assist with editing parts of the text (Sections~\ref{sec:intro}--\ref{sec:expe}); all AI-assisted text was reviewed and verified by the authors, who take full responsibility for the paper.
%The authors used Claude (Claude Opus 5.5, Anthropic)~\cite{anthropic2026claude} to assist with editing and revising parts of the manuscript text, specifically in Sections~\ref{sec:intro}, \ref{sec:related}, Sections~\ref{sec:method} and~\ref{sec:expe}. All AI-assisted text was reviewed, verified and edited by the authors, who take full responsibility for the content of this paper.

\bibliographystyle{IEEEtran}
\bibliography{IJCNN2027_submission/egbib}

\end{document}